\documentclass[conference]{IEEEtran}

\usepackage[hidelinks]{hyperref}
\usepackage{doi}
\usepackage{graphicx}
\usepackage{changes}

\title{Learning abstract perceptual notions: the example of space}
\author{\IEEEauthorblockN{Alexander V. Terekhov}
\IEEEauthorblockA{Laboratoire Psychologie de la Perception\\Universit\'{e} Paris Descartes\\75006 Paris, France\\
Email: avterekhov@gmail.com}
\and
\IEEEauthorblockN{J. Kevin O'Regan}
\IEEEauthorblockA{Laboratoire Psychologie de la Perception\\Universit\'{e} Paris Descartes\\75006 Paris, France\\
Email: jkevin.oregan@gmail.com}
}
\begin{document}

\maketitle
\begin{abstract}
Humans are extremely swift learners. We are able to grasp highly abstract notions, whether they come from art perception or pure mathematics. Current machine learning techniques demonstrate astonishing results in extracting patterns in information. Yet the abstract notions we possess are more than just statistical patterns in the incoming information. Sensorimotor theory suggests that they represent functions, laws, describing how the information can be transformed, or, in other words, they represent the statistics of sensorimotor changes rather than sensory inputs themselves. The aim of our work is to suggest a way for machine learning and sensorimotor theory to benefit from each other so as to pave the way toward new horizons in learning. We show in this study that a highly abstract notion, that of space, can be seen as a collection of laws of transformations of sensory information and that these laws could in theory be learned by a naive agent. As an illustration we do a one-dimensional simulation in which an agent extracts spatial knowledge in the form of internalized (``sensible'') rigid displacements. The agent uses them to encode its own displacements in a way which is isometrically related to external space. Though the algorithm allowing acquisition of rigid displacements is designed \emph{ad hoc}, we believe it can stimulate the development of unsupervised learning techniques leading to similar results.
\end{abstract}

\section{Introduction}
Humans are highly capable of extracting regularities in the information they receive, be it simple sensory experience, like images or sounds, or highly abstract notions like structure of texts or mathematical concepts. The fact that this ability is not restricted to any particular type of information suggests that it is assured by a mechanism ubiquitously present in the brain at all levels of information processing. Identifying this mechanism is a Holy Graal of artificial intelligence and computational neuroscience.

During the last decade significant progress has been made in this direction in the processing of sensory information. Autoencoding neural networks have been shown to extract basic statistical regularities in their inputs, and when applied for example to natural images, such networks can develop receptive fields similar to those present in the human visual system~\cite{Olshausen1996,Schmidhuber1996,Lee2008}. Cascades of autoencoders, called deep belief networks, have been shown to identify previously unspecified patterns~\cite{Hinton2006}, such as the presence of kittens of any kind in their input~\cite{Wired2012}, and to reflect highly abstract notions, such as giving an account of the numerosity of distributed objects in the input~\cite{Stoianov2012}.

Yet in spite of the evident step forward offered by deep belief networks, their limitations are becoming more apparent. These networks are mainly used to learn features, that is, invariant patterns in their inputs. They are not generally used to learn the laws of how these patterns \emph{transform} by motor commands. For example, making a neural network capable of predicting the shift of an image corresponding to the robot's actual motion is known to be challenging. The same task becomes trivial when approached by making use of sensorimotor transformations~\cite{Censi2012}. Such an ability to predict the consequences of one's own actions, known as a forward model, plays a central role in human perception~\cite{Clark2013}. More generally, humans are also able to extract abstract notions characterized not only by grouping of less abstract notions, but also by identifying similar functions, affordances, and laws of transformation present in grouped notions.

The sensorimotor approach to perception~\cite{ORegan2001} emphasizes the central role of information transformations in perception, as well as in higher cognitive notions as `self' and `consciousness'~\cite{ORegan2011}. For example, the perception of the body, as seen from the sensorimotor standpoint, constitutes a collection of knowledge of how different sensory stimulations can be evoked or altered by one's own actions. Thus, the feeling of a touch on my arm is constituted by all the changes that I know will occur when I move my arm (if I move it away, sensation will cease; if I move it closer to the touch, sensation will increase...), or when I look towards the touched location (I will see something touching my arm), etc. This approach though providing an interesting conceptual basis to the study of artificial perception, does not give any concrete suggestions bringing nearer implementation of perceptual systems in artificial agents.

It is necessary to bring the sensorimotor approach closer to the point where it could be combined with existing machine learning solutions, such as those offered by deep belief networks. As a step in this direction one may look into how a naive artificial agent could learn a familiar perceptual notion, which would be essentially sensorimotor, and hence inaccessible for standard pattern-extracting techniques. We decided on the notion of space as particularly suited for this purpose. 

\subsection{Space}
\begin{figure*}[!t]
\centering
\includegraphics[width=\linewidth]{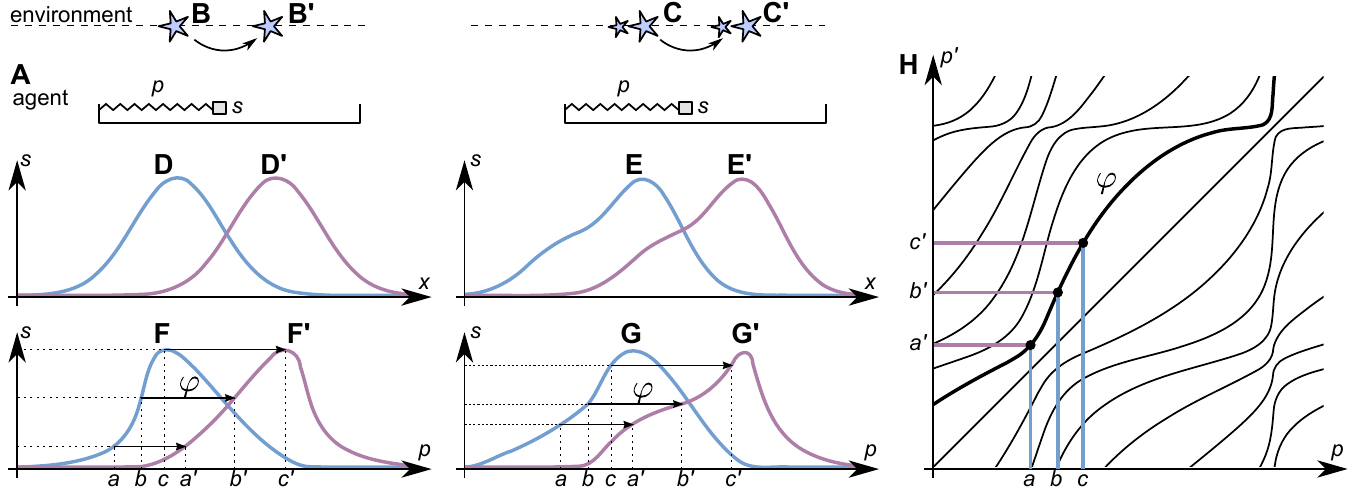}
\caption{Algorithm of space acquisition illustrated with a simplified agent. The agent (A) has the form of a tray, inside which a photoreceptor $s$ moves with the help of a muscle, scanning the environment (B) composed of scattered light sources. The length of the muscle is linked to the output of the proprioceptive cell $p$ in a systematic, but unknown way. The output of the photoreceptor depends on its position $x$ in real space (D). The agent learns the sensorimotor contingency (F) linking $p$ and $s$. After a rigid displacement of the agent, or a corresponding displacement of the environment from B to B', the output of the photoreceptor changes from D to D' and a new sensorimotor contingency F' is established. For a sufficiently small rigid displacement the outputs of the photoreceptor will overlap before and after the displacement. The agent makes a record of the corresponding proprioceptive values between the sensorimotor contingency F and F' (arrows from $a$,$b$,$c$ to $a'$,$b'$,$c'$) and constructs the function $p' = \varphi(p)$ (H, bold line). Different functions $\varphi$ (thin lines in H) correspond to different rigid displacements. If the agent faces a different environment C and makes a rigid displacement equivalent to its displacement to C', the outputs of the photoreceptor change from E to E' and the corresponding sensorimotor contingency changes from G to G'. Yet the same function $\varphi$ links G to G'.}
\label{fig_sim}
\end{figure*}
The notion of space is inherently amodal: it is present in all our perceptual modalities. We usually attribute our visual, auditory, tactile and even olfactory~\cite{Wachowiak2011} experience to certain domains of space. Similarly we think of our body as a collection of objects occupying certain spatial volumes, which we can fairly well localize due to kinesthesia. We see from this that the notion of space is not derived from the sensory information per se. None of our sensory modalities directly measures ``spatiality''. Instead, space represents a framework within which we interpret incoming sensory inputs. These aspects of space were emphasized by Poincar\'{e} and Nicod who suggested that our perception of space is not attributed to particular sensory inputs, but rather to regularities present in those inputs. In particular Poincar\'{e} stated that if our world was not mainly made of rigid objects we would never be able to develop spatial perception. 

The problem of the acquisition of spatial or geometrical notions has been addressed in a number of studies in robotics~\cite{Pierce1997,Bowling2007,Stober2011}, computational neuroscience~\cite{Wyss2006} as well as in developmental studies~\cite{McGurk1974,Slater1990,Spencer2001,Newcombe2003,Cheng2005,Lourenco2005,Izard2009a}. The specificity of the current study is however that it does not aim at understanding the mechanisms of space acquisition. Instead we aim at approaching the general principles that lead to the emergence of highly abstract notions, which are not simply based on pure statistics on the sensory inputs, but reflect the regularities in the way these inputs change. We take space just as an example of such an abstract notion.

We will consider a naive agent that has very limited prior knowledge about itself and about the environment in which it is immersed. It does not know that space is a manifold, even less that it is Euclidean. In this respect the current study is different from similar works~\cite{Philipona2003,Laflaquiere2010,Laflaquiere2012} which made use of the implicit assumption that space has the structure of a vector space.

\section{Conceptual scheme}

Consider the sensory universe or ``Merkwelt'' (cf von Uexk\"ull \cite{Uexkull1957}) of the one-dimensional agent in Fig 1. Assume (though this is not known to the agent) that its body is composed of a single photoreceptive sensor \(s\) that can move laterally inside its body using a ``muscle'' (Fig 1A). Assume a one-dimensional environment as in Fig 1B, and assume first that it is static. If the agent were to perform scanning actions with the muscle and were to plot photoreceptor output against the photoreceptor's actual physical position \(x\), it would obtain a plot such as Fig 1D. But it cannot do this because it has no notion, let alone any measure, of physical position, and only has knowledge of proprioception \(p\). The agent can only plot photoreceptor output against proprioception, and so obtains a distorted plot as in Fig. 1F. This ``sensorimotor contingency''\cite{Mackay1962,ORegan2001} is all that the agent knows about. It does not know anything about the structure of its body and sensor, let alone that there is such a thing as space in which it is immersed. Indeed the agent does not need such notions to understand its world, since its world is completely accounted for by its knowledge of the sensorimotor contingency it has established by scanning.

But now suppose that the environment can move relative to the agent, for example taking Fig 1B to Fig 1B'. The previously plotted sensorimotor contingency will no longer apply, and a different plot will be obtained (e.g. Fig 1F'). The agent goes from being able to completely predict the effects of its scanning actions on its sensory input, to no longer being able to do so.

However, there is a notable fact which applies. Although the agent does not know this, physicists looking from outside the agent would note that if the displacement relative to the environment is not too large, there will be some overlap between the physical locations scanned before and after the displacement. In this overlapping region, the sensor occupies the same positions relative to the environment as it occupied before the displacement occurred. Since sensory input depends only on the position of the photoreceptor  relative  to the environment, the agent will thus discover that for these positions the sensory input from the photoreceptor will be the same as before the displacement.

Registering such a coincidence is not uncommon for an agent with a single photoreceptor, but the same would happen for an agent with numerous receptors. For such a more complicated agent the coincidence would be extremely noteworthy.

In an attempt to better ``understand'' its environment, the agent will thus naturally make a catalogue of these coincidences (cf. arrows in Fig 1F, F'), and so establish a function \(\varphi\) linking the values of proprioception observed before a change to the corresponding values of proprioception after the change. Such a function for all values of proprioception is shown in Fig 1H.

Assume that over time, the environment displaces rigidly to various extents, with the agent located initially at various positions. Furthermore, assume that such displacements can happen for entirely different environments (e.g. Fig 1C). Since the sensorimotor contingencies themselves depend on all these factors, it might be expected that different functions \(\varphi\) would have to be catalogued for all these different cases. Yet it is a remarkable fact that the set of functions \(\varphi\) is much simpler: for a given displacement of the environment, the agent will discover the \emph{same} functions \(\varphi\), even when this displacement starts from different initial positions, and even when the environment is different. 

We shall see below that this remarkable simplicity of the functions \(\varphi\) \emph{provides the agent with the notion of space}. But first let us see where the simplicity derives from.

Each function \(\varphi\) links proprioceptive values before an environmental change to proprioceptive values after the change, in such a way that for the linked values the outputs of the photoreceptor match before and after the change. Seen from outside the agent, the physicist would know that this situation will occur if the agent's photoreceptor occupies the same position relative to the environment before and after the environmental change. And this will happen \emph{if (1) the environment makes a rigid displacement, and if (2) the agent's photoreceptor makes a rigid displacement equal to the rigid displacement of the environment}. Thus physicists looking at the agent would know that the functions \(\varphi\) actually measure, in proprioceptive coordinates, rigid physical displacements of the environment relative to the agent (or vice versa).

Now we can understand why the set of functions phi is so simple: it is because a defining property of rigid displacements is that they are independent of their starting points, and independent of the properties of what is being displaced.

The functions \(\varphi\) can thus be seen as perceptual constructs equivalent to physical rigid displacements, or one could say, following Jean Nicod\cite{Nicod1923}, that they are \emph{sensible rigid displacements}, where \emph{sensible} refers to the fact that they are \emph{defined within the Merkwelt of the agent}.

\section{Simulations}

\subsection{Agent}

\begin{figure}[!t]
\centering
\includegraphics[width=\linewidth]{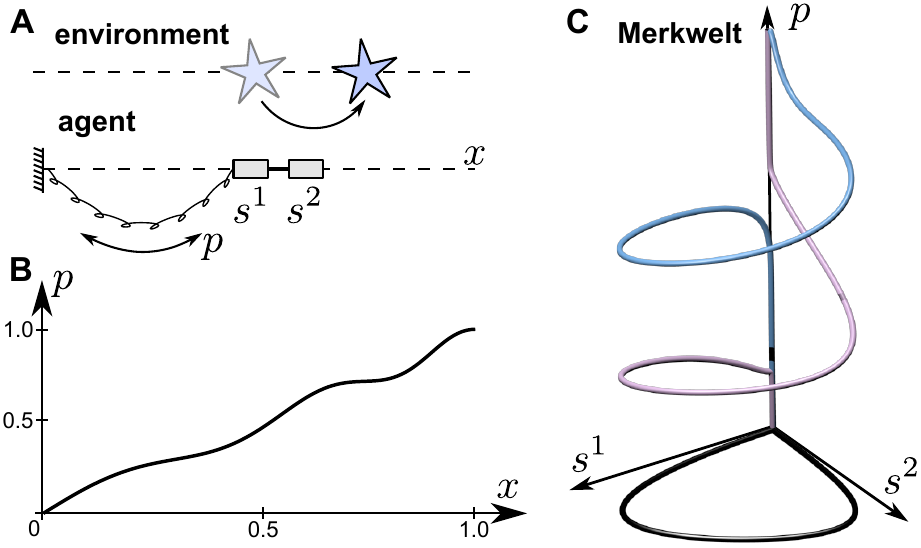}
\caption{Simulated one-dimensional agent. A. The agent consists of two rigidly connected photoreceptive cells, \(s^1\) and \(s^2\), articulated by a muscle, whose elongation is measured by a proprioceptive cell \(p\). The proprioception encodes the position of the photoreceptive cells, though in a highly non-linear way, illustrated in B. The agent is exposed to an environment filled with light sources (a star in A), which it scans by moving the photoreceptors and tabulating the values \(p, s^1, s^2\). These values can be illustrated in the agent's Merkwelt as a single curved (pink curve in C). If the environment displaces, the same scanning movements result in a different curve (blue curve in C). Yet because the outputs of the photoreceptors depend on their relative position with respect to the light-source, there will be coinciding values of \(s^1\) and \(s^2\) before and after the environment displacement. Thus, the two curves have the same projection on the \((s_1, s_2)\) plane denoted with a black line in C.}
\label{fig2}
\end{figure}

We used a computer simulation to illustrate the feasibility of the described procedure. The agent was similar to the one presented in Figure~1A with the only difference that it had two rigidly connected photoreceptive cells \(s = (s^1,s^2)^T\) placed at 0.1 distance from each other (see Figure~2A). Having multiple photoreceptive cells decreases the probability of a situation when the same exteroceptory input can be associated with several distinct proprioceptive states. Thus having multiple rigidly connected exteroceptors makes the estimation of the functions \(\varphi\) more robust. We decided on having just two photoreceptive cells and a single proprioceptive input because in this case the agent's Merkwelt is three-dimensional and can still be illustrated. Each photoreceptive cell had a Gaussian tuning curve with respect to the relative position of the photoreceptor and the light source. The response was maximal (it was equal to one) when the light-source was in front of the receptor and decreased as a Gaussian function with 0.1 variance as the light-source shifted sideways. In case of several light sources the summed excitation was computed.

We took a highly non-linear relationship between single-valued proprioception \(p\) and spatial coordinates of the middle point between proprioceptors \(x\). This relationship is illustrated in Figure~2B. The environment was composed of several light-sources placed at a distance from the line of agent's displacements. The agent's Merkwelt for a single light source environment is illustrated in Figure~2C. Each curve in Figure~2C corresponds to a different relative position between the environment and the agent. Note that both curves have the same projection on the \((s^1,s^2)\) plane, which corresponds to the coincidences of the exteroceptory inputs.

In the simulated experiment the agent was programmed to perform scanning motions by displacing its exteroceptors between the minimal \(p_{min} = 0\) and maximal \(p_{max} = 1\) admissible values of proprioception, which corresponded to the spatial locations \(x=0\) and \(x=1\) of the mid-point between photoreceptors. The proprioception \(p\) took values \(p_{min} = p_1, p_2, \cdots, p_N = p_{max}\) changing with a constant step \(0.001\). The agent was exposed to an environment composed of five light sources placed randomly in the \([-4,5]\) region. The agent collected the tuples  \(\langle p_i, s^1_i, s^2_i\rangle\), which were assumed to be veridical for the first scan. The values from further scans were compared to previously collected tuples (see below).

\subsection{Acquisition of \(\varphi\)'s}

\begin{figure*}[!t]
\centering
\includegraphics[width=\linewidth]{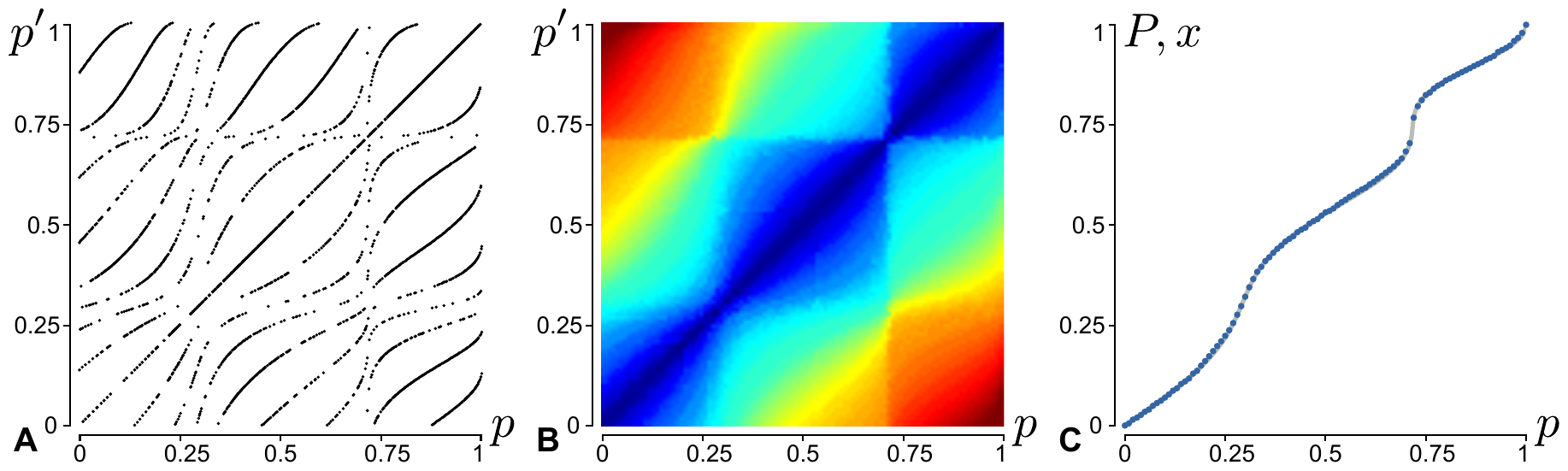}
\caption{Sensible rigid displacements underlying space. A. The functions \(\varphi\) corresponding to different rigid displacements as learned by the agent. The pointwise interruptions in the functions occur close to singular points in the mapping between the displacement of the photoreceptors and the corresponding changes in proprioception. B. The functions \(\varphi\) can be used to define a metric on proprioception, where the distance between two values \(p_1\) and \(p_2\) is defined as a norm of the function \(\varphi\) mapping one into another (see text). This metric is denoted with color code. C. The new metric defines a new parametrization \(P\) of proprioception \(p\). The correspondence between \(p\) and \(P\) (blue dotted line) closely matches the correspondence between \(p\) and the physical position of the photoreceptors \(x\) (gray line).}
\label{fig3}
\end{figure*}

The agent was unexpectedly displaced to a random extent proportional to 0.01 and not exceeding 0.95 in magnitude, thus ensuring at least 5\% overlap with the previous scanning. The agent detected the displacement by the deviations it caused between the incoming sensory inputs and the stored tuples. The agent was programmed to classify the new scan as corresponding to a new environment if the deviation in any of the photoreceptors exceeded 0.05. The agent then collected and stored a new set of tuples \(\langle p'_i, s'_i{}^1, s_i'{}^2\rangle\). The function \(\varphi\) corresponding to the displacement of the environment was determined as a set of pairs \(\{\langle p_i, p'_i\rangle\}\) for which the norm \(\|s_i-s_i'\|\) did not exceed 0.2\% of the maximal value of \(s_1\) and \(s_2\) across the scan.

The described procedure was repeated 191 times. The resultant functions \(\varphi\) are presented in Figure~3A. Note the interrupted character of the functions \(\varphi\). The interruptions are most pronounced around the values 0.3 or 0.7 either on the \(p\) or the \(p'\) axis. These values correspond to the points of vanishing derivative of the mapping between the actual physical position of the receptors and their proprioceptive coordinates. At these points a large displacement of the photoreceptors provokes only a minute change in proprioception, so explaining an interruption in the functions \(\varphi\). Note that we deliberately made the environment of the agent sufficiently rich, so that the closeness of the photoreceptors values would imply closeness of the relative spatial location between the environment and the sensors. We believe that this requirement can be relaxed by allowing the agent to scan different environments and keeping only those functions \(\varphi\) that work for all of them. We will address this issue in future work.

\subsection{Rescaling proprioception}

The functions \(\varphi\) presented in Figure~3A define the agent's sensible rigid displacements, and these now can be used to solve various spatial tasks. As an illustration of the close relationship between the functions \(\varphi\) and space we show how they can be used to rescale proprioception to linearize the highly non-linear correspondence between spatial location of the photoreceptors and the outputs of proprioception.

The rescaling implies producing a new metric space of variables \(P\) out of the existing metric space of \(p\)'s by changing the metric of the latter. The current metric on \(p\)'s is simply
\[
\rho(p_1, p_2) = |p_1 - p_2|.
\]

We define a new metric \(\mu\) based on the functions \(\varphi\):
\[
\mu(p_1, p_2) = \|\varphi_{p_1\to p_2}\|,
\]
where \(\varphi_{p_1\to p_2}\) is such that
\[
p_2 = \varphi_{p_1\to p_2}(p_1),
\]
and \(\|\cdot\|\) is a norm of \(\varphi\), such that the identity function has a zero norm. Here we use
\[
\|\varphi\| = \max_{p}|p-\varphi(p)|.
\]

Note that because of the properties of the functions \(\varphi\) the function \(\mu\) is a metric. The symmetry condition holds because of the symmetry of the functions \(\varphi\) with respect to the diagonal line in Figure~3A. The first condition holds because for equal \(p_1\) and \(p_2\) the function \(\varphi_{p_1\to p_2}\) is the identity function and its norm is zero. The triangular inequality holds simply because \(\|\cdot\|\) is a norm on \(\varphi\)'s.

In simulations we took a regular sampling of proprioception with step 0.01 and computed a distance matrix for every pair of values from the sampling. The corresponding distance is presented in Figure 3B. Theoretically this matrix defines the values of \(\mu\) and hence it defines the rescaled proprioception \(P\). For the sake of illustration we applied multidimensional scaling (\texttt{STRESS1}) to the distance matrix. The resultant embedding was then shifted and normalized to simplify the comparison with the old values \(p\).

Figure 3C presents the values \(P\) of rescaled proprioception plotted against the original values \(p\) as well as the values \(x\) of the true spatial location of the photoreceptors. One can notice the similarity between values \(P\) and \(x\), which means that the new proprioception \(P\) is linearly related to \(x\).

\section{Conclusion}

The present article has taken the specific example of space, and showed that the basic structure of space can be discovered by looking at the laws of change of the sensory information under rigid displacements of the environment. This is expected from the central idea of the sensorimotor theory, namely the idea that abstract notions do not reflect regularities in the sensory inputs per se. Instead, they reflect robust laws describing the \emph{possible changes} of sensory inputs following action on the part of the agent. The crucial difference between the former and the latter is that the laws are grounded in the regularities of the sensorimotor changes, rather than in sensations themselves.

As mentioned in the introduction, we see our -- sensorimotor -- approach as complementary to classical unsupervised learning techniques such as deep learning. Making the interface with such techniques is one of the primary goals for our future work. In the short term we see deep belief networks as a possible tool for learning sensorimotor laws, such as those represented by the functions \(\varphi\). In the simplest case, the functions \(\varphi\) could be fitted by neural networks trained in a supervised manner to match the coincidences in the sensory inputs.

Although we have not done so here, our ultimate purpose would now be to find a way of automatically discovering these laws (in the case of space, the \(\varphi\) functions).  A guideline for the search for such a principle could be the idea that all biological agents strive to account for their sensorimotor information in the most economical way. The functions \(\varphi\) provide such an economical description in the case of space, because they define rigid displacements, which can be applied to any possible environment. Moreover, these functions allow the agent to separate the displacement of the environment, such as due to its own displacements and hence not deserving special attention, from other changes, caused by external factors and thus worth noting.

Finding a principle that would allow the emergence of the functions \(\varphi\) or similar, and implementing this principle using machine learning techniques would be a way of approaching our goal of melding together ideas from machine learning and ideas from sensorimotor theory. The link will be productive, we think, because whereas machine learning provides efficient techniques for capturing statistical properties of the incoming information, what is needed in order to extract higher level abstract constructs is  the laws of \emph{transformation} of sensory information.

The present article has only made a first step in this direction, by taking the specific example of space and an oversimplified agent, and showing what the laws are that must be extracted, without however as yet describing how this could be done autonomously. 

It must be noted that though we illustrate our approach with an oversimplified example only, the approach can be straightforwardly extended to more complicated agents. An example for a two-dimensional agent can be found in the preprint~\cite{Terekhov2013}. A nice property of our approach is that the complexity of the underlying computations depends only weakly on the number of sensors. For example, we could use millions of camera pixels instead of the pair of photoreceptors without complicating the computations significantly. The same can be said about proprioception: substituting a single measuring unit with an ensemble of them would not appreciably complicate the computation. The increase of dimensionality of the space inhabited by the agent will, however, inevitably require a larger amount of data and might require reducing the precision in order to allow implementation on state-of-the-art robotic systems. But as developmental studies show, the notion of space remains immature until the age of 6-10 years~\cite{Dillon2013}, suggesting that a large amount of data is necessary in order to develop spatial perception.

\section*{Acknowledgement} This work was funded by the European Research Council (FP 7 Program) ERC Advanced Grant ``FEEL'' to KO'R

\bibliographystyle{IEEEtran}
\bibliography{biblio}

\begin{thebibliography}{10}
\providecommand{\url}[1]{#1}
\csname url@samestyle\endcsname
\providecommand{\newblock}{\relax}
\providecommand{\bibinfo}[2]{#2}
\providecommand{\BIBentrySTDinterwordspacing}{\spaceskip=0pt\relax}
\providecommand{\BIBentryALTinterwordstretchfactor}{4}
\providecommand{\BIBentryALTinterwordspacing}{\spaceskip=\fontdimen2\font plus
\BIBentryALTinterwordstretchfactor\fontdimen3\font minus
  \fontdimen4\font\relax}
\providecommand{\BIBforeignlanguage}[2]{{%
\expandafter\ifx\csname l@#1\endcsname\relax
\typeout{** WARNING: IEEEtran.bst: No hyphenation pattern has been}%
\typeout{** loaded for the language `#1'. Using the pattern for}%
\typeout{** the default language instead.}%
\else
\language=\csname l@#1\endcsname
\fi
#2}}
\providecommand{\BIBdecl}{\relax}
\BIBdecl

\bibitem{Olshausen1996}
\BIBentryALTinterwordspacing
B.~A. Olshausen and D.~J. Field, ``\BIBforeignlanguage{eng}{Emergence of
  simple-cell receptive field properties by learning a sparse code for natural
  images.}'' \emph{\BIBforeignlanguage{eng}{Nature}}, vol. 381, no. 6583, pp.
  607--609, Jun 1996. [Online]. Available:
  \url{http://dx.doi.org/10.1038/381607a0}
\BIBentrySTDinterwordspacing

\bibitem{Schmidhuber1996}
J.~Schmidhuber, M.~Eldracher, and B.~Foltin, ``Semilinear predictability
  minimization produces well-known feature detectors,'' \emph{Neural
  Computation}, vol.~8, pp. 773--786, 1996.

\bibitem{Lee2008}
H.~Lee, C.~Ekanadham, and A.~Ng, ``Sparse deep belief net model for visual area
  v2,'' in \emph{Advances in Neural Information Processing Systems 20}, J.~C.
  Platt, D.~Koller, Y.~Singer, and S.~Roweis, Eds.\hskip 1em plus 0.5em minus
  0.4em\relax Cambridge, MA: MIT Press, 2008, pp. 873--880.

\bibitem{Hinton2006}
\BIBentryALTinterwordspacing
G.~E. Hinton and R.~R. Salakhutdinov, ``\BIBforeignlanguage{eng}{Reducing the
  dimensionality of data with neural networks.}''
  \emph{\BIBforeignlanguage{eng}{Science}}, vol. 313, no. 5786, pp. 504--507,
  Jul 2006. [Online]. Available:
  \url{http://dx.doi.org/10.1126/science.1127647}
\BIBentrySTDinterwordspacing

\bibitem{Wired2012}
\BIBentryALTinterwordspacing
C.~Liat. (2012, June) \BIBforeignlanguage{English}{Google brain simulator
  identifies cats on youtube}. wired.co.uk. [Online]. Available:
  \url{http://www.wired.co.uk/news/archive/2012-06/26/google-brain-recognises-cats}
\BIBentrySTDinterwordspacing

\bibitem{Stoianov2012}
\BIBentryALTinterwordspacing
I.~Stoianov and M.~Zorzi, ``\BIBforeignlanguage{eng}{Emergence of a 'visual
  number sense' in hierarchical generative models.}''
  \emph{\BIBforeignlanguage{eng}{Nat Neurosci}}, vol.~15, no.~2, pp. 194--196,
  Feb 2012. [Online]. Available: \url{http://dx.doi.org/10.1038/nn.2996}
\BIBentrySTDinterwordspacing

\bibitem{Censi2012}
A.~Censi and R.~Murray, ``Learning diffeomorphism models of robotic
  sensorimotor cascades,'' in \emph{Robotics and Automation (ICRA), 2012 IEEE
  International Conference on}, May 2012, pp. 3657--3664.

\bibitem{Clark2013}
\BIBentryALTinterwordspacing
A.~Clark, ``\BIBforeignlanguage{eng}{Whatever next? predictive brains, situated
  agents, and the future of cognitive science.}''
  \emph{\BIBforeignlanguage{eng}{Behav Brain Sci}}, vol.~36, no.~3, pp.
  181--204, Jun 2013. [Online]. Available:
  \url{http://dx.doi.org/10.1017/S0140525X12000477}
\BIBentrySTDinterwordspacing

\bibitem{ORegan2001}
J.~K. O'Regan and A.~Noë, ``\BIBforeignlanguage{eng}{A sensorimotor account of
  vision and visual consciousness.}'' \emph{\BIBforeignlanguage{eng}{Behav
  Brain Sci}}, vol.~24, no.~5, pp. 939--73; discussion 973--1031, Oct 2001.

\bibitem{ORegan2011}
J.~K. O'Regan, \emph{Why Red Doesn't Sound Like a Bell: Understanding the feel
  of consciousness}.\hskip 1em plus 0.5em minus 0.4em\relax Oxford University
  Press, 2011.

\bibitem{Wachowiak2011}
\BIBentryALTinterwordspacing
M.~Wachowiak, ``\BIBforeignlanguage{eng}{All in a sniff: olfaction as a model
  for active sensing.}'' \emph{\BIBforeignlanguage{eng}{Neuron}}, vol.~71,
  no.~6, pp. 962--973, Sep 2011. [Online]. Available:
  \url{http://dx.doi.org/10.1016/j.neuron.2011.08.030}
\BIBentrySTDinterwordspacing

\bibitem{Pierce1997}
\BIBentryALTinterwordspacing
D.~Pierce and B.~J. Kuipers, ``Map learning with uninterpreted sensors and
  effectors,'' \emph{Artificial Intelligence}, vol.~92, no. 1–2, pp. 169 --
  227, 1997. [Online]. Available:
  \url{http://www.sciencedirect.com/science/article/pii/S0004370296000513}
\BIBentrySTDinterwordspacing

\bibitem{Bowling2007}
\BIBentryALTinterwordspacing
M.~Bowling, D.~Wilkinson, A.~Ghodsi, and A.~Milstein,
  ``\BIBforeignlanguage{English}{Subjective localization with action respecting
  embedding},'' in \emph{\BIBforeignlanguage{English}{Robotics Research}}, ser.
  Springer Tracts in Advanced Robotics, S.~Thrun, R.~Brooks, and
  H.~Durrant-Whyte, Eds.\hskip 1em plus 0.5em minus 0.4em\relax Springer Berlin
  Heidelberg, 2007, vol.~28, pp. 190--202. [Online]. Available:
  \url{http://dx.doi.org/10.1007/978-3-540-48113-3_18}
\BIBentrySTDinterwordspacing

\bibitem{Stober2011}
J.~Stober, R.~Miikkulainen, and B.~Kuipers, ``Learning geometry from
  sensorimotor experience,'' in \emph{Development and Learning (ICDL), 2011
  IEEE International Conference on}, vol.~2, 2011, pp. 1--6.

\bibitem{Wyss2006}
\BIBentryALTinterwordspacing
R.~Wyss, P.~K{\"{o}}nig, and P.~F. M.~J. Verschure,
  ``\BIBforeignlanguage{eng}{A model of the ventral visual system based on
  temporal stability and local memory.}'' \emph{\BIBforeignlanguage{eng}{PLoS
  Biol}}, vol.~4, no.~5, p. e120, May 2006. [Online]. Available:
  \url{http://dx.doi.org/10.1371/journal.pbio.0040120}
\BIBentrySTDinterwordspacing

\bibitem{McGurk1974}
H.~McGurk and M.~Lewis, ``\BIBforeignlanguage{eng}{Space perception in early
  infancy: perception within a common auditory-visual space?}''
  \emph{\BIBforeignlanguage{eng}{Science}}, vol. 186, no. 4164, pp. 649--650,
  Nov 1974.

\bibitem{Slater1990}
A.~Slater, A.~Mattock, and E.~Brown, ``\BIBforeignlanguage{eng}{Size constancy
  at birth: newborn infants' responses to retinal and real size.}''
  \emph{\BIBforeignlanguage{eng}{J Exp Child Psychol}}, vol.~49, no.~2, pp.
  314--322, Apr 1990.

\bibitem{Spencer2001}
J.~P. Spencer, L.~B. Smith, and E.~Thelen, ``\BIBforeignlanguage{eng}{Tests of
  a dynamic systems account of the a-not-b error: the influence of prior
  experience on the spatial memory abilities of two-year-olds.}''
  \emph{\BIBforeignlanguage{eng}{Child Dev}}, vol.~72, no.~5, pp. 1327--1346,
  2001.

\bibitem{Newcombe2003}
N.~S. Newcombe and J.~Huttenlocher, \emph{Making space: The development of
  spatial representation and reasoning}.\hskip 1em plus 0.5em minus 0.4em\relax
  MIT Press, 2003.

\bibitem{Cheng2005}
\BIBentryALTinterwordspacing
K.~Cheng and N.~Newcombe, ``\BIBforeignlanguage{English}{Is there a geometric
  module for spatial orientation? squaring theory and evidence},''
  \emph{\BIBforeignlanguage{English}{Psychonomic Bulletin \& Review}}, vol.~12,
  no.~1, pp. 1--23, 2005. [Online]. Available:
  \url{http://dx.doi.org/10.3758/BF03196346}
\BIBentrySTDinterwordspacing

\bibitem{Lourenco2005}
\BIBentryALTinterwordspacing
S.~F. Lourenco, J.~Huttenlocher, and M.~Vasilyeva,
  ``\BIBforeignlanguage{eng}{Toddlers' representations of space: the role of
  viewer perspective.}'' \emph{\BIBforeignlanguage{eng}{Psychol Sci}}, vol.~16,
  no.~4, pp. 255--259, Apr 2005. [Online]. Available:
  \url{http://dx.doi.org/10.1111/j.0956-7976.2005.01524.x}
\BIBentrySTDinterwordspacing

\bibitem{Izard2009a}
V.~Izard and E.~S. Spelke, ``\BIBforeignlanguage{eng}{Development of
  sensitivity to geometry in visual forms.}''
  \emph{\BIBforeignlanguage{eng}{Hum Evol}}, vol.~23, no.~3, pp. 213--248, Jan
  2009.

\bibitem{Philipona2003}
\BIBentryALTinterwordspacing
D.~Philipona, J.~K. O'Regan, and J.-P. Nadal, ``\BIBforeignlanguage{eng}{Is
  there something out there? inferring space from sensorimotor dependencies.}''
  \emph{\BIBforeignlanguage{eng}{Neural Comput}}, vol.~15, no.~9, pp.
  2029--2049, Sep 2003. [Online]. Available:
  \url{http://dx.doi.org/10.1162/089976603322297278}
\BIBentrySTDinterwordspacing

\bibitem{Laflaquiere2010}
A.~Laflaquiere, S.~Argentieri, B.~Gas, and E.~Castillo-Castenada, ``Space
  dimension perception from the multimodal sensorimotor flow of a naive robotic
  agent,'' in \emph{Intelligent Robots and Systems (IROS), 2010 IEEE/RSJ
  International Conference on}.\hskip 1em plus 0.5em minus 0.4em\relax IEEE,
  2010, pp. 1520--1525.

\bibitem{Laflaquiere2012}
A.~Laflaquiere, S.~Argentieri, O.~Breysse, S.~Genet, and B.~Gas, ``A non-linear
  approach to space dimension perception by a naive agent,'' in
  \emph{Intelligent Robots and Systems (IROS), 2012 IEEE/RSJ International
  Conference on}.\hskip 1em plus 0.5em minus 0.4em\relax IEEE, 2012, pp.
  3253--3259.

\bibitem{Uexkull1957}
J.~von Uexküll, ``A stroll through the worlds of animals and men: A picture
  book of invisible worlds,'' in \emph{Instinctive Behavior: The Development of
  a Modern Concept}, C.~H. Schiller, Ed.\hskip 1em plus 0.5em minus 0.4em\relax
  New York: International Universities Press, Inc., 1957, pp. 5--80.

\bibitem{Mackay1962}
D.~M. MacKay, ``Theoretical models of space perception,'' \emph{Aspects of the
  theory of artificial intelligence}, pp. 83--104, 1962.

\bibitem{Nicod1923}
J.~Nicod, \emph{Foundations of Geometry and Induction}, reprint~ed.\hskip 1em
  plus 0.5em minus 0.4em\relax Routledge, 2000.

\bibitem{Terekhov2013}
A.~V. Terekhov and J.~K. O'Regan, ``Space as an invention of biological
  organisms,'' \emph{arXiv}, p. 1308.2124, 2013.

\bibitem{Dillon2013}
\BIBentryALTinterwordspacing
M.~R. Dillon, Y.~Huang, and E.~S. Spelke, ``\BIBforeignlanguage{eng}{Core
  foundations of abstract geometry.}'' \emph{\BIBforeignlanguage{eng}{Proc Natl
  Acad Sci U S A}}, vol. 110, no.~35, pp. 14\,191--14\,195, Aug 2013. [Online].
  Available: \url{http://dx.doi.org/10.1073/pnas.1312640110}
\BIBentrySTDinterwordspacing

\end{thebibliography}

\end{document}